# *Formalization of Psychological Knowledge in Answer Set Programming and its Application*

Marcello Balduccini

*Intelligent Systems, KRL*
*Eastman Kodak Company*
*Rochester, NY 14650-2102 USA*
`marcello.balduccini@gmail.com`

Sara Girotto

*Department of Psychology*
*Texas Tech University*
*Lubbock, TX 79409 USA*
`sara.girotto@ttu.edu`

## Abstract

In this paper we explore the use of Answer Set Programming (ASP) to formalize, and reason about, psychological knowledge. In the field of psychology, a considerable amount of knowledge is still expressed using only natural language. This lack of a formalization complicates accurate studies, comparisons, and verification of theories. We believe that ASP, a knowledge representation formalism allowing for concise and simple representation of defaults, uncertainty, and evolving domains, can be used successfully for the formalization of psychological knowledge. To demonstrate the viability of ASP for this task, in this paper we develop an ASP-based formalization of the mechanics of Short-Term Memory. We also show that our approach can have rather immediate practical uses by demonstrating an application of our formalization to the task of predicting a user's interaction with a graphical interface.

## 1 Introduction

In this paper we explore the use of Answer Set Programming (ASP) (Gelfond and Lifschitz 1991; Marek and Truszczynski 1999) to formalize psychological knowledge and to reason about it. The importance of a precise formalization of scientific knowledge has been known for a long time (see e.g. Hilbert's philosophy of physics). Most notably, formalizing a body of knowledge in an area improves one's ability to (1) accurately study the properties and consequences of sets of statements, (2) compare competing sets of statements, and (3) design experiments aimed at confirming or refuting sets of statements. In the field of psychology, some of the theories about the mechanisms that govern the brain have been formalized using artificial neural networks and similar tools (e.g. (McCarley et al. 2002)).

That approach works well for theories that can be expressed in quantitative terms. However, theories of a more qualitative or logical nature, which by their own nature do not provide precise quantitative predictions, are not easy to formalize in this way. We believe that ASP can be used successfully for the formalization of such bodies of knowledge. ASP is a knowledge representation formalism allowing for concise and simple representation of defaults, uncertainty, and evolving domains, and has been demonstrated to be a useful paradigm for the formalization of knowledge of various kinds (e.g. (Baral and Gelfond 2005; Son and Sakama 2009)). One further advantage of ASP is that it is directly executable, in the sense that the consequences of collections of ASP statements can be directly, and often efficiently, computed using computer programs.

To demonstrate the viability of ASP for the formalization of psychological knowledge, in this paper we develop an ASP-based formalization of the mechanics of Short-Term Memory (STM) and chunking. We selected this theory because it is rather mature and, like the general type of psychological knowledge that we aim to formalize, it is mostly of a qualitative nature, and is expressed, in the literature, at a rather high level of abstraction. Moreover, formalizing it is challenging, because it involves modeling a sophisticated dynamic domain involving non-determinism, fixed-capacity storage, and time-based decay. *The combination of these features makes it rather difficult to use other languages for the encoding.* As a confirmation of the benefits of formalizing psychological knowledge, we show that our formalization immediately allows us to point out an element of the model of STM that may deserve further investigation. We also show that our approach can have rather direct practical uses by demonstrating an application of our formalization to the task of predicting a user's interaction with a graphical interface. To the best of our knowledge, ours is the first attempt to use ASP for the formalization of theories about the brain, although it is worth mentioning that ASP has been previously applied previously to the specification of a user's mental model for recommender systems (Leite and Ilic 2007). Although for our formalization we follow a rather well-known theory of STM and chunking (e.g. (Kassin 2006; Nevid 2007)), we do not intend to claim this to be the "correct" theory. On the contrary, any objections to the theory that our formalization may raise are a further demonstration of the benefits of formalizing psychological knowledge.

The paper is organized as follows. We start with some background on ASP and on the representation of dynamic domains. Next, we provide an account of the mechanics of STM and chunking, as it is commonly found in psychology literature. Then, we describe our ASP-based formalization of the mechanics of STM and chunking, and show how the formalization can be used for the task of predicting a user's interaction with a graphical interface. Finally, we conclude with a brief discussion on what we have achieved and on possible extensions.

## 2 Answer Set Programming and Dynamic Domains

Let us begin by giving some background on ASP. We define the syntax of the language precisely, but only give the informal semantics of the language in order to save space. We refer the reader to (Gelfond and Lifschitz 1991; Niemela and Simons 2000) for a specification of the formal semantics. Let $\Sigma$ be a signature containing constant, function and predicate symbols. Terms and atoms are formed as usual in first-order logic. A (basic) lit-eral is either an atom $a$ or its strong (also called classical or epistemic) negation $\neg a$. A *rule* is a statement of the form:

$$h_1 \vee \ldots \vee h_k \leftarrow l_1, \ldots, l_m, \text{not } l_{m+1}, \ldots, \text{not } l_n$$

where $h_i$'s and $l_i$'s are ground literals and *not* is the so-called *default negation*. The intuitive meaning of the rule is that a reasoner who believes $\{l_1, \ldots, l_m\}$ and has no reason to believe $\{l_{m+1}, \ldots, l_n\}$, must believe one of $h_i$'s. Symbol $\leftarrow$ can be omitted if no $l_i$'s are specified. Often, rules of the form $h \leftarrow \text{not } h, l_1, \ldots, \text{not } l_n$ are abbreviated into $\leftarrow l_1, \ldots, \text{not } l_n$, and called *constraints*. The intuitive meaning of a constraint is that $\{l_1, \ldots, l_m, \text{not } l_{m+1}, \ldots, \text{not } l_n\}$ must not be satisfied. A rule containing variables is interpreted as the shorthand for the set of rules obtained by replacing the variables with all the possible ground terms. A *program* is a pair $\langle \Sigma, \Pi \rangle$, where $\Sigma$ is a signature and $\Pi$ is a set of rules over $\Sigma$. We often denote programs just by the second element of the pair, and let the signature be defined implicitly. Finally, the *answer set* (or *model*) of a program $\Pi$ is the collection of its consequences under the answer set semantics. Notice that the semantics of ASP is defined in such a way that programs may have multiple answer sets, intuitively corresponding to alternative views of the specification given by the program. In that respect, the semantics of default negation provides a simple way of encoding choices. For example, the set of rules $\{p \leftarrow \text{not } q.\ q \leftarrow \text{not } p.\}$ intuitively states that either $p$ or $q$ hold, and the corresponding program has two answer sets, $\{p\}$, $\{q\}$. Because a convenient representation of alternatives is often important in the formalization of knowledge, the language of ASP has been extended with *constraint literals* (Niemela and Simons 2000), which are expressions of the form $m\{l_1, l_2, \ldots, l_k\}n$, where $m$, $n$ are arithmetic expressions and $l_i$'s are basic literals as defined above. A constraint literal is satisfied whenever the number of literals that hold from $\{l_1, \ldots, l_k\}$ is between $m$ and $n$, inclusive. Using constraint literals, the choice between $p$ and $q$, under some set of conditions $\Gamma$, can be compactly encoded by the rule $1\{p, q\}1 \leftarrow \Gamma$. A rule of this form is called *choice rule*. To further increase flexibility, the set $\{l_1, \ldots, l_k\}$ can also be specified as $\{l(\vec{X}) : d(\vec{X})\}$, where $\vec{X}$ is a list of variables. Such an expression intuitively stands for the set of all $l(\vec{x})$ such that $d(\vec{x})$ holds. We refer the reader to (Niemela and Simons 2000) for a more detailed definition of the syntax of constraint literals and of the corresponding extended rules.

Because of the dynamic nature of STM, for its formalization we use techniques from the area of the representation of dynamic (or evolving) domains. The key elements of the representation techniques are presented next; we refer the readers to e.g. (Gelfond and Lifschitz 1998; Gelfond 2002) for more details. *Fluents* are first-order ground terms, and intuitively denote the properties of interest of the domain (whose truth value typically depends upon time). For example, an expression of the form $on(block_1, block_2)$ is a fluent, and may mean that $block_1$ is on top of $block_2$. A fluent literal is either a fluent $f$ or its negation ($\neg f$). Actions are also first-order ground terms. For example, $move(block_3, block_2)$ may mean that $block_3$ is moved on top of $block_2$. A set of fluent literals is *consistent* if, for every fluent $f$, $f$ and $\neg f$ do not both belong to the set. A set of fluent literals is *complete* if, for every fluent $f$, either $f$ or $\neg f$ belong to the set. The set of all the possible evolutions of a dynamic domain is represented by a *transition diagram*, i.e. a directed graph whose nodes – each labeled by a consistent set of fluent literals – correspond to the states of the domain in which the properties specified are respectively true or false, and whose arcs – each labeled by a set of actions – correspond to the occurrence of state transitions due to the execution of the actions specified. When complete knowledge is available about a state, the corresponding set of fluent literals is also complete. When instead a set of fluent literals is not complete, that means that the knowledge about the corresponding state is incomplete (e.g. it is unknown whether $f$ or $\neg f$ holds). Incomplete or partial states are typically used to represent uncertainty about domains.

Because the size of transition diagrams grows exponentially with the increase of the number of properties of the domain and actions, a direct representation is usually impractical. Instead, transition diagrams are encoded using an indirect representation, based on the research on action languages (Gelfond and Lifschitz 1998). Because we are not aware of any well-established action language that combines all the features needed for our model, and because in this paper the focus is not on developing a new action language, here we adopt the variant of writing such encoding directly in ASP – see e.g. (Balduccini et al. 2000). In fact, because of the high level of abstraction of ASP, encoding knowledge (even about dynamic domains) directly in ASP rather than in action languages is nowadays common practice (see e.g. (Delgrande et al. 2009; Thielscher 2009) for some recent examples).

The encoding is based on the notion of a path in the transition diagram from a given initial state, corresponding to a particular possible evolution of the domain from that initial state. The steps in a path are identified by integers (with 0 denoting the initial state), and logical statements (often called *laws*) are used to encode, in general terms, the transitions from one step to the next. The fact that a fluent $f$ holds at a step $i$ in the evolution of the domain is represented by the expression $h(f, i)$, where relation $h$ stands for *holds*. If $\neg f$ is true, we write $\neg h(f, i)$. Occurrences of actions are represented by expressions of the form $o(a, i)$, saying that action $a$ occurs at step $i$ ($o$ stands for *occurs*). An *action description* is a collection of laws describing the evolution of the domain. Given an action description $AD$, a description of the initial state $\sigma_0$ (e.g. $\sigma_0 = \{h(f_1, 0), \neg h(f_2, 0), \ldots\}$), and a sequence of occurrences of actions $\alpha$ (e.g. $\alpha = \{o(a_1, 0), o(a_3, 0), o(a_4, 1), \ldots\}$), the corresponding path(s) in the transition diagram can be computed by finding the answer set(s) of $AD \cup \sigma_0 \cup \alpha$.

### 3 Short-Term Memory

Short-Term Memory is "the memory storage system that allows for short-term retention of information before it is either transferred to long-term memory or forgotten" (Nevid 2007). This view is based on the so called *three-stage model of memory* (Atkinson and Shiffrin 1971): sensory inputs are first stored in Sensory Memory, which is very volatile and has large capacity; then, a portion of the inputs is processed – and possibly transformed into more rich representations – and moved to Short-Term Memory, which is less volatile than Sensory Memory, but of limited capacity. Short-Term Memory is also often viewed as a working memory, i.e. as a location where information is processed (Card et al. 1983). Finally, selected information is moved to Long-Term Memory, which has larger capacity and longer retention periods.

Beginning in the 1950s, several studies have been conducted to determine the capacity of STM. Miller (Miller 1956) reported evidence showing that the capacity of STM in humans is of 7 pieces of information. Later studies have lowered the capacity limit of STM to about 4 pieces of information (e.g. (Cowan 2000)). Interestingly, the limit on the number of pieces of information that STM can hold does not affect directly the *amount* of information (in an information-theoretic sense) that STM can hold. In fact, STM appears to be capable to storing *references* to concepts that are stored in Long-Term Memory. Although one such reference counts as a single piece of information toward the capacity limit of STM, the amount of information it conveys can be large. For example, it has been observed that it is normally difficult for people to remember the 12-letter sequence CN NIB MMT VU SA, while most people have no problems remembering the sequence CNN IBM MTV USA, because each triplet refers to a concept stored in Long-Term Memory, and can thus be represented in STM by just 4 symbols (Kassin 2006). The phenomenon of the detection and use of known patterns in STM is referred to as *chunking*.

Another limit of STM is that the information it contains is retained only for a short period of time, often set to about 30 seconds by researchers ((Nevid 2007); notice however that the issue of a time limit on the information stored in STM is somewhat controversial – see e.g. (Cowan 2000; Card et al. 1983)).[1] This limit can be extended by performing *maintenance rehearsal*, which consists in consciously repeating over and over the information that needs to be preserved. To increase the flexibility of our formalization, in the next section we abstract from specific values for the limits of capacity and retention over time, and rather write our model in a parametric way. This makes it possible, among other things, to use our formalization to analyze the effects of different choices for these parameters, effectively allowing us to compare variants of the theory of STM.

### 4 A Formalization of Short-Term Memory

Using a common methodology in ASP-based knowledge representation, we begin our discussion on the formalization by condensing the description of STM and chunking in a number of statements still written in natural language, but precisely formulated. Later, we will encode those statements using ASP. The statements are: (1) STM is a collection of symbols; (2) The size of STM is limited to $\omega$ elements; (3) Each symbol has an expiration time associated with it (saying when the piece of information will be "forgotten"); (4) Symbols are divided into primitive symbols and chunk symbols; (5) Primitive symbols represent concepts that are innate in the brain (e.g. are direct encoding of sensory input); (6) Chunk symbols represent concepts that are stored in Long-Term Memory; (7) New symbols can be added to STM. If a symbol is added to STM when $\omega$ elements are already in STM, the symbol that is closest to expiring is removed from STM ("forgotten"). In the case of multiple symbols equally close to expiring, one is selected arbitrarily; (8) When a symbol is added to STM, or when a symbol from STM is used (in particular, when performing maintenance rehearsal), its expiration time is reset to a constant value $\varepsilon$; (9) *Simplifying assumption:* only a single operation (where by operation we mean either addition or use) can occur on STM at any given time. The next set of statements describes the mechanism of chunking: (10) A chunk is a set of (primitive or chunk) symbols; (11) A chunk symbol is a symbol that uniquely denotes a chunk; (12) A chunk is detected in STM if all the symbols that it consists of are in STM; (13) When a chunk is detected in STM, the symbols it consists of are removed from STM and the corresponding chunk symbol is added to STM; (14) A symbol can be inferred from STM if it belongs to STM, or if it is part of a chunk that can be inferred from STM; [2] (15) *Simplifying assumption:* chunks can only be detected when STM is not in use (i.e. no addition or use operations are being performed); (16) *Simplifying assumption:* at every step, at most one chunk can be detected.

---

[1] For example, according to (Card et al. 1983), decay is affected by variables such as the number of chunks that the user is trying to remember, and retrieval interference with similar chunks.

Now we are ready to focus on the formalization of STM and chunking in ASP. Fluent $in\_stm(s)$ says that symbol $s$ (where $s$ is a possibly compound term) is in STM; $expiration(s, k)$ says that symbol $s$ will expire (i.e. will be "forgotten") in $k$ units of time, unless the expiration counter is otherwise altered. Action $store(s)$ says that symbol $s$ is stored in STM; $use(s)$ says that $s$ is used (typically, during maintenance rehearsal). Relation $primitive(s)$ says that $s$ is a primitive symbol; chunks are described by relation $chunk(s)$, saying that $s$ is a chunk symbol, and by relation $chunk\_element(s, s')$, stating that $s'$ is a component of the chunk identified by $s$; relation $symbol(s)$ says that $s$ is a symbol (either primitive or chunk). Relation $stm\_max\_size(\omega)$ says that the size of STM is limited to $\omega$ elements; $stm\_expiration(\varepsilon)$ states that the symbols in STM expire after $\varepsilon$ units of time. Finally, in order to update the expiration counters based on the duration of the actions executed at each step, relation $dur(i, d)$ says that the overall duration of step $i$, based on the actions that took place, is $d$ units of time.

We divide the axioms in our formalization of the STM[3] in a number of categories. We begin with the axioms that describe the effect of storing a symbol $s$ in STM. The first axiom says that an effect of storing $s$ in STM is that $s$ becomes part of STM:

$$h(in\_stm(S), I + 1) \leftarrow o(store(S), I).$$

The next axiom says that, if adding a symbol to STM would cause the STM size limit to be exceeded, then the symbol that is closest to expiring will be forgotten. Notice that the simplifying assumptions listed among the natural language statements above guarantee that there is a unique such symbol, thus simplifying the writing of the axiom (lifting the assumptions is easy, but would lengthen the presentation).

$$\neg h(in\_stm(S2), I + 1) \leftarrow$$
$$\quad S1 \neq S2,\ o(store(S1), I),\ stm\_max\_size(MX),$$
$$\quad curr\_stm\_size(MX, I),\ not\ some\_expiring(I),\ oldest\_in\_stm(S2, I).$$

The axiom relies on a number of auxiliary relations. These relations depend on the current step in the evolution of the domain, and hence could be represented as fluents. However, to stress the distinction between "regular" fluents and auxiliary relations, we prefer the alternative writing that does not rely upon relation $h$, thus writing e.g. $curr\_stm\_size(MX, I)$ instead of $h(curr\_stm\_size(MX), I)$. Relation $curr\_stm\_size(w, i)$ says that the size of STM at step $i$ is $w$; $some\_expiring(i)$ states that some symbol in STM will be forgotten

---

[2] Notice that this statement can be applied recursively.

[3] To save space, we omit most atoms formed by domain predicates and, in a few rules, use default negation directly instead of writing a separate rule for closed-world assumption. For example, if $p$ holds whenever $q$ is false and $q$ is assumed to be false unless it is known to be true, we might write $p \leftarrow not\ q$ instead of the more methodologically correct $\{p \leftarrow \neg q.\ \neg q \leftarrow not\ q.\}$.

Formalization of Psychological Knowledge in ASP and its Application 7when the domain moves to the next step; $oldest\_in\_stm(s, i)$ says that $s$ is the symbol closest to expiring. The accurate reading of the above axiom is then: if $s_1$ is stored in STM at step $i$, and STM already contains $\omega$ symbols, none of which are due to expire at the next time step, then the oldest symbol in STM will no longer be in STM at step $i + 1$. The auxiliary relations are defined as follows:

$$oldest\_in\_stm(S, I) \leftarrow \\ \quad h(expiration(S, E), I), \\ \quad not\ smaller\_expiration(E, I).$$

$$smaller\_expiration(E1, I) \leftarrow \\ \quad h(expiration(S, E2), I), \\ \quad E2 < E1.$$

$$curr\_stm\_size(N, I) \leftarrow \\ \quad N\ \{\ h(in\_stm(S), I) : symbol(S)\ \}\ N.$$

$$expiring(S, I) \leftarrow \\ \quad h(expiration(S, SK), I), \\ \quad dur(I, D), SK \leq D.$$

$$some\_expiring(I) \leftarrow \\ \quad expiring(S, I).$$

The next axiom states that an effect of storing $s$ in STM is that the expiration time of $s$ is set to $\varepsilon$:

$$h(expiration(S, E), I + 1) \leftarrow stm\_expiration(E),\ o(store(S), I).$$

The next axiom (together with some auxiliary definitions) says that it is impossible for two STM-related actions to be executed at the same time.

$$\leftarrow o(A1, I), o(A2, I),\ A1 \neq A2,\ stm\_related(A1),\ stm\_related(A2).$$
$$stm\_related(store(S)) \leftarrow symbol(S).$$
$$stm\_related(use(S)) \leftarrow symbol(S).$$

The final axiom in this category states that, whenever a symbol from STM is used, its expiration time is reset.

$$h(expiration(S, E), I + 1) \leftarrow \\ \quad stm\_expiration(E), \\ \quad o(use(S), I), \\ \quad h(in\_stm(S), I).$$

The second group of axioms describes the mechanism of chunking. Three axioms state that, when the components of a chunk are detected in STM, the corresponding symbols are replaced by the chunk symbol[4], whose expiration is set to $\varepsilon$.

$$\neg h(in\_stm(S), I+1) \leftarrow$$
$$\quad detected(C, I),$$
$$\quad chunk\_element(C, S).$$
$$h(in\_stm(C), I+1) \leftarrow$$
$$\quad detected(C, I).$$
$$h(expiration(C, E), I+1) \leftarrow$$
$$\quad stm\_expiration(E),$$
$$\quad detected(C, I).$$

Auxiliary relation $detected(C, I)$, used above, says that chunk $C$ has been detected in STM at step $I$. The detection occurs, when STM is not in use as per our simplifying assumption above, by checking if there is any chunk whose components are all in STM. If symbols corresponding to multiple chunks are available in STM, only one chunk is detected at every step, as per assumption (16) on page 6. The choice of which chunk is detected is non-deterministic, and encoded using a choice rule[5], as follows:

$$1 \{ detected(C, I) : \neg chunk\_element\_missing(C, I) \} 1 \leftarrow$$
$$\quad stm\_idle(I),$$
$$\quad chunk\_detectable(I).$$
$$chunk\_detectable(I) \leftarrow$$
$$\quad step(I),$$
$$\quad chunk(C),$$
$$\quad stm\_idle(I),$$
$$\quad \neg chunk\_element\_missing(C, I).$$
$$\neg chunk\_element\_missing(C, I) \leftarrow$$
$$\quad chunk(C),$$
$$\quad \text{not } chunk\_element\_missing(C, I).$$
$$chunk\_element\_missing(C, I) \leftarrow$$
$$\quad chunk\_element(C, S),$$
$$\quad \neg h(in\_stm(S), I).$$
$$\neg stm\_idle(I) \leftarrow$$
$$\quad o(A, I),$$
$$\quad memory\_related(A).$$
$$stm\_idle(I) \leftarrow$$
$$\quad \text{not } \neg stm\_idle(I).$$

It is interesting to note that the components of a detected chunk are allowed to be located anywhere in STM. However, *now that the model is formalized at this level of detail, one cannot help but wonder whether in reality the focus of the mechanism of chunking is on symbols that have been added more recently.* We were unable to find published studies regarding this issue.

The final group of axioms deals with the evolution of the contents of STM over time,

---

[4]  Our simplifying assumption that at most one chunk can be detected at every step ensures that the number of items in STM does not increase as a result of the chunking process.

[5] Readers who are familiar with ASP may notice that we allow the use of non-domain predicate $\neg chunk\_element\_missing(C, I)$ in the head of the choice rule. This is done to keep the presentation short. From the perspective of the implementation, when using ASP parsers that expect a domain predicate, one would have to use a slightly longer encoding.

both when *store* or *use* actions occur, and when they do not. Notice that action theories often assume that fluents maintain their truth value *by inertia* unless they are forced to change by the occurrence of actions. In the case of fluents $in\_stm(s)$ and $expiration(s, k)$, however, the evolution over time is more complex (and such fluents are then called *non-inertial*). In fact, for every symbol $s$ in STM, $expiration(s, \varepsilon)$ holds at first, but then $expiration(s, \varepsilon)$ becomes false and $expiration(s, \varepsilon - \delta)$ becomes true, where $\delta$ is the duration of the latest step, and so on. On the other hand, $in\_stm(s)$ is true if-and-only-if $expiration(s, k)$ holds for some $k > 0$. The following axioms accomplish three tasks: they define the behavior of inertial fluents, using a rather standard ASP encoding of the inertia axiom, which relies on ASP's ability to encode defaults (the rule for $\neg h(F, I+1)$, omitted, is similar); they state that fluents $in\_stm(s)$ and $expiration(s, k)$ are non-inertial (the reason for doing so explicitly will become clear in the next section); and formalize the default evolution of the non-inertial fluents' truth value over time. Notice that relation $expiring$, used here, was defined above (page 7).

$$h(F, I+1) \leftarrow$$
$$\quad h(F, I),$$
$$\quad \text{not } \neg h(F, I+1),$$
$$\quad \text{not } noninertial(F).$$
$$noninertial(in\_stm(S)) \leftarrow$$
$$\quad symbol(S).$$
$$noninertial(expiration(S, E)) \leftarrow$$
$$\quad symbol(S),$$
$$\quad expiration\_value(E).$$
$$h(in\_stm(S), I+1) \leftarrow$$
$$\quad h(in\_stm(S), I),$$
$$\quad \text{not } expiring(S, I),$$
$$\quad \text{not } \neg h(in\_stm(S), I+1).$$
$$\neg h(in\_stm(S), I+1) \leftarrow$$
$$\quad h(in\_stm(S), I),$$
$$\quad expiring(S, I),$$
$$\quad \text{not } h(in\_stm(S), I+1).$$
$$\neg h(in\_stm(S), I+1) \leftarrow$$
$$\quad \neg h(in\_stm(S), I),$$
$$\quad \text{not } h(in\_stm(S), I+1).$$
$$h(expiration(S, E-D), I+1) \leftarrow$$
$$\quad expiration\_value(E),$$
$$\quad h(expiration(S, E), I),$$
$$\quad dur(I, D), E > D,$$
$$\quad \text{not } different\_expiration(S, E-D, I+1),$$
$$\quad \text{not } \neg h(in\_stm(S), I+1).$$
$$different\_expiration(S, E1, I) \leftarrow$$
$$\quad expiration\_value(E1),$$
$$\quad expiration\_value(E2),$$
$$\quad E2 \neq E1,$$
$$\quad h(expiration(S, E2), I).$$

To demonstrate that our formalization captures the key features of the mechanics of STM and chunking, we subject it to a test of memory span. In a memory-span test, a subject is presented with a sequence of digits, and is asked to reproduce the sequence (Kassin 2006). By increasing the length of the sequence and by allowing or avoiding the occurrence of familiar sub-sequences of digits, one can verify the capacity limit of STM and the role of chunking. Because here we are only concerned with correctly modeling STM, we abstract from the way digits are actually read, and rather represent the acquisition of the sequence of digits directly as the occurrence of suitable $store(s)$ actions. Similarly, the final reproduction of the sequence is replaced by checking the contents of STM at the end of the experiment. As common in ASP, all computations are reduced to finding answer sets of suitable programs, and the results of the experiments are determined by observing the values of the relevant fluents in such answer sets.

From now on, we refer to the above formalization of STM by $\Pi_{STM}$. Boundary conditions that are shared by all the instances of the memory-span test are encoded by the set $\Pi_P$ of rules, shown below. The first two rules of $\Pi_P$ set the value of $\omega$ to a capacity of 4 symbols (in line with (Cowan 2000)) and the value of $\varepsilon$ to 30 time units. The next rule states that each step has a duration of 1 time unit. This set-up intuitively corresponds to a scenario in which STM has a 30 second time limit on the retention of information and the digits are presented at a rate of one per second. The last three rules define the set of primitives for the memory-span test. We use the expression $seq(p, d)$ to represent the fact that digit $d$ is at position $p$ in the sequence.

$$stm\_max\_size(4).$$
$$stm\_expiration(30).$$
$$dur(I, 1).$$
$$position(1).\ position(2).\ \ldots\ position(6).$$
$$digit(0).\ digit(1).\ \ldots\ digit(9).$$
$$primitive(seq(P, D)) \leftarrow position(P), digit(D).$$

The initial state of STM is such that no symbols are initially in STM. This is encoded by $\sigma_{STM}$:

$$\neg h(in\_stm(S), 0) \leftarrow symbol(S).$$

In the first instance, STM is presented with the sequence $2, 4, 5, 7$. Human subjects are normally able to reproduce this sequence. Let us see if our formalization can do the same. The sequence of digits is encoded by set $SPAN_1$ of rules:

$$o(store(seq(1, 2)), 0).\ o(store(seq(2, 4)), 1).$$
$$o(store(seq(3, 5)), 2).\ o(store(seq(4, 7)), 3).$$

To predict the behavior of STM and determine which symbols will be in it at the end of the experiment, we need to look at the path in the transition diagram from the initial state, described by $\sigma_{STM}$, and under the occurrence of the actions in $SPAN_1$. As explained earlier in this paper, this can be accomplished by finding the answer set of $\Pi_1 = \Pi_{STM} \cup \Pi_P \cup \sigma_{STM} \cup SPAN_1$. It is not difficult to check that, at step 4 (corresponding to the end of the experiment), the state of STM is:

$$h(in\_stm(seq(1, 2)), 4),\ h(in\_stm(seq(2, 4)), 4),$$
$$h(in\_stm(seq(3, 5)), 4),\ h(in\_stm(seq(4, 7)), 4),$$
$$h(expiration(seq(1, 2), 27), 4),\ h(expiration(seq(2, 4), 28), 4),$$
$$h(expiration(seq(3, 5), 29), 4),\ h(expiration(seq(4, 7), 30), 4),$$

which shows that the sequence is remembered correctly (and far from being forgotten, as the expiration times show). Let us now consider another instance, in which the sequence of digits is $2, 4, 5, 7, 3$. The corresponding $store(s)$ actions are encoded by $SPAN_2 = SPAN_1 \cup \{o(store(seq(5,3)),4)\}$. This sequence is beyond the capacity of STM stated in $\Pi_P$. Human subjects are unable to reproduce sequences beyond the capacity of STM (unless chunking occurs). Our formalization of STM exhibits the same behavior. In fact, according to the answer set of program $\Pi_2 = \Pi_{STM} \cup \Pi_P \cup \sigma_{STM} \cup SPAN_2$ the state of STM at the end of the experiment is:

$$h(in\_stm(seq(2,4)),5), \ h(in\_stm(seq(3,5)),5),$$
$$h(in\_stm(seq(4,7)),5), \ h(in\_stm(seq(5,3)),5).$$

As expected, the first element of the sequence has been forgotten. In the next instance, we consider the sequence $5, 8, 5 - 8, 0, 6$, where "$-$" represents a 1-second pause in the presentation of the digits. This sequence is, in principle, beyond the capacity of STM, but we further assume familiarity with the area codes $585$ and $806$. Under these conditions, human subjects have demonstrated to be capable of remembering the sequence after noticing the presence of the area codes in them.[6] The knowledge about the area codes is encoded by the set $\Gamma_1$ of rules:

$$chunk(seq(P, ac(roc))).$$
$$chunk\_element(seq(P, ac(roc)), seq(P, 5)).$$
$$chunk\_element(seq(P, ac(roc)), seq(P+1, 8)).$$
$$chunk\_element(seq(P, ac(roc)), seq(P+2, 5)).$$

$$chunk(seq(P, ac(lbb))).$$
$$chunk\_element(seq(P, ac(lbb)), seq(P, 8)).$$
$$chunk\_element(seq(P, ac(lbb)), seq(P+1, 0)).$$
$$chunk\_element(seq(P, ac(lbb)), seq(P+2, 6)).$$

The sequence of digits is encoded by $SPAN_3$:

$$o(store(seq(1,5)),0). \ o(store(seq(2,8)),1).$$
$$o(store(seq(3,5)),2). \ \% \text{ no action at step } 3$$
$$o(store(seq(4,8)),4). \ o(store(seq(5,0)),5).$$
$$o(store(seq(6,6)),6).$$

At the end of the experiment (we select step $8$ to allow sufficient time for the chunking of the second triplet to occur), the state of STM predicted by our formalization is:

$$h(in\_stm(seq(1, ac(roc))), 8),$$
$$h(in\_stm(seq(4, ac(lbb))), 8),$$
$$h(expiration(seq(1, ac(roc)), 26), 8),$$
$$h(expiration(seq(4, ac(lbb)), 30), 8)$$

which shows that the chunking of the two area codes has occurred, allowing STM to store a sequence of digits that is longer than $\omega$ symbols. In the final instance of this section, we consider two additional chunks, $5, 8, 5, 2$ (supposedly a pin number), and $1, 3$ (considered an unlucky number in various cultures), which are encoded with the same technique shown above. The corresponding set of rules $\Gamma'_1$ is obtained from $\Gamma_1$ by adding the encoding of the new chunks, called $my\_pin$ and $unlucky\_13$. The sequence to be remembered is $5, 8, 5 - 2, 1, 3$, encoded by $SPAN_4$:

$$o(store(seq(1,5)), 0).\ o(store(seq(2,8)), 1).$$
$$o(store(seq(3,5)), 2).\ \%\ \text{no action at step 3}$$
$$o(store(seq(4,2)), 4).\ o(store(seq(5,1)), 5).$$
$$o(store(seq(6,3)), 6).$$

---

[6] The 1-second pause is used to allow sufficient time for the detection of the first chunk. Subject studies have shown that chunk detection does not occur or occurs with difficulty when the stimuli are presented at too high a frequency.

Notice that, at step 7, chunks $my\_pin$ and $unlucky\_13$ will both be available for detection. The literature does not specify any particular order in which the chunks are detected by the brain, and thus one should assume that the order of detection is arbitrary.[7] It is not difficult to show that our formalization correctly yields *two answer sets*, both predicting a final state of STM in which $my\_pin$ and $unlucky\_13$ are in STM, but differing for the order which chunking occurs. One answer set, in fact, predicts the chunking of $my\_pin$ first: $\{detected(seq(1, my\_pin), 7), detected(seq(5, unlucky\_13), 8)\}$, while the other predicts the chunking of $my\_pin$ last: $\{detected(seq(5, unlucky\_13), 7),\ detected(seq(1, my\_pin), 8)\}$. Although space restrictions prevent us from formalizing alternative theories of STM and perform an analytical comparison, it is worth noting that even the single formalization developed allows comparing (similar) variants of the theory corresponding to different values of $\omega$ and $\varepsilon$. One could for example repeat the above experiments with different parameter values and compare the predicted behavior with actual subject behavior, thus confirming or refuting some of those variants.

## 5 A Practical Application

The availability of a formalization of psychological knowledge not only allows better analysis, comparison, and verification of psychological theories, but may also have more immediate practical applications. In this section we show how our formalization can be used to predict a user's difficulties in interacting with a graphical user interface. Psychological theories used in human-computer interaction (e.g. (Card et al. 1983; Kieras and Polson 1983; Kieras and Polson 1985)) for this kind of evaluations are often of a qualitative nature and do not allow for precise quantitative predictions. The ability to accurately encode this type of knowledge in ASP allows one to use psychological theories to draw accurate conclusions, and to do so automatically. We believe this to be a clear step forward from the use of rules of thumb and guidelines common nowadays in human-computer interaction. We consider a scenario in which the user is told a sequence of tasks (menu clicks, tab selections, etc.) to be performed, and is expected to execute it without being reminded about any task. If the user succeeds, then that means that the sequence can be stored completely in STM (possibly chunked), and that the sequence is also short enough not to be forgotten during execution. This scenario corresponds for example to a situation in which a user is trying to follow the instructions on a help page. To begin, we need to formalize the sequence of tasks and its execution. Basic operations are $click(m)$, corresponding to clicking menu item $m$; $select(t)$, meaning that tab $t$ in a dialog box is selected; $check(c)$, corresponding to putting a checkmark in checkbox $c$. The primitive STM symbols are of the form $task(n, op)$, where $op$ is one of the basic operations, and $n$ is its index in the sequence. The sequence is stored in STM by means of occurrences of $store(s)$ actions, as before. We use term $task(n, op)$ also as one of the fluents of the formalization. Its meaning is that, based on the state of the reasoner's memory, $op$ is the $n^{th}$ task in the sequence. The fluent is non-inertial, because it is defined directly by the current state of STM:

$$noninertial(task(N, A)).$$
$$h(task(N, A), I) \leftarrow stm\_h(task(N, A), I).$$
$$stm\_h(S, I) \leftarrow h(in\_stm(S), I).$$
$$stm\_h(S, I) \leftarrow stm\_h(C, I), chunk\_element(C, S).$$

---

[7] Simplifying assumption (16) prohibits concurrent detection.

From the point of view of knowledge representation, it is worth noting how ASP makes it easy to express the recursive definition of relation $stm\_h(s, i)$. The other key fluents of the formalization of the sequence are $current\_task(n)$, which says that the current task is the $n^{th}$ in the sequence, $task\_forgotten(n)$, meaning that the $n^{th}$ task has been forgotten, and $completed$, meaning that the sequence has been completed. Task number 1 is selected as current task as soon as it becomes available (the use of defaults here and later greatly shortens the representation):

$$h(current\_task(1), I) \leftarrow has\_task(1, I), \text{not } \neg h(current\_task(1), I).$$

Relation $has\_task(n, i)$ (definition omitted to save space) says that, at step $i$, the sequence has a task with index $n$. Performing the current task makes the following task become current, unless the task that was performed was the last in the sequence, in which case the sequence is complete:

$$h(current\_task(N+1), I+1) \leftarrow$$
$$\quad o(A, I), h(current\_task(N), I),$$
$$\quad h(task(N, A), I), \text{not } last\_task(N, I).$$
$$h(completed, I+1) \leftarrow$$
$$\quad o(A, T), h(current\_task(N), I),$$
$$\quad h(task(N, A), I), last\_task(N, I).$$

(Other rules that ensure that only one task is current at any time are omitted.) The reasoner detects that the current task has been forgotten when it cannot recall it:

$$h(task\_forgotten(N), I) \leftarrow h(current\_task(N), I), \text{not } has\_task(N, I).$$

Task execution is encoded by a rule saying that, if the current task is executable, then the reasoner will perform it:

$$o(A, T) \leftarrow h(current\_task(N), I), h(task(N, A), I),$$
$$\quad current\_task\_executable(I).$$
$$current\_task\_executable(I) \leftarrow$$
$$\quad h(current\_task(N), I), h(task(N, A), I), \text{not } \neg o(A, I).$$

Finally, we make the duration of the user's actions depend on his skill level (the rules are straightforward and omitted to save space). For expert users, every action takes 1 unit of time. For beginner users, who need to scan the screen to look for the items to act upon, clicking a menu takes 3 units; clicking a sub-menu takes 5; selecting a tab in a dialog, as well as putting a checkmark in a checkbox, takes 7.[8] We denote the above set of rules by $\Pi_{TASK}$. To test our formalization, we consider an example in which a *beginner* user is given the following sequence, $UI_1$. (STM-related parameters are $\omega = 4$ and $\varepsilon = 30$ as before.)

$$o(store(task(1, click(m(tools)))), 0).$$
$$o(store(task(2, click(subm(options)))), 1).$$
$$o(store(task(3, select(tab(text)))), 3).$$
$$o(store(task(4, check(ck(use\_hardtabs)))), 4).$$
$$o(store(task(5, select(tab(highlighting)))), 6).$$
$$o(store(task(6, check(ck(detect\_language)))), 7).$$
$$-o(A, I) \leftarrow external\_action(A), I \leq 8.$$

Notice the pauses at steps 2, 5, 8, to allow enough time for chunking to occur. The last rule ensures that the reasoner waits to have acquired the complete sequence before beginning to execute it. The user is assumed to be familiar with the combinations "click tools, click options," and "select tab 'text', put a checkmark on 'use hard-tabs'." The encoding, $\Gamma_2^a$, of the first chunk is:

$chunk(task(N, tools\_options)).$
$chunk\_element(task(N, tools\_options), task(N, click(m(tools)))).$
$chunk\_element(task(N, tools\_options), task(N+1, click(subm(options)))).$

The second chunk is encoded similarly. The encoding of the two chunks is denoted by $\Gamma_2$. Because of our assumption that beginner users take some time to locate the graphical items, one can expect that the user will forget the sequence before completing it. The use of our formalization allows one to make the prediction more precise. In fact, the answer set of the program $\Pi_1^{UI} = \Pi_{STM} \cup \Pi_{TASK} \cup \sigma_{STM} \cup UI_1 \cup \Gamma_2$ shows that: (1) chunking occurs and the sequence fits in STM, as shown in the answer set by atoms such as:

$h(in\_stm(task(1, tools\_options)), 12),$
$h(in\_stm(task(3, text\_htabs)), 12),$
$h(in\_stm(task(5, select(tab(highlighting)))), 12),$
$h(in\_stm(task(6, check(ck(detect\_language)))), 12).$

However, the answer set also shows that (2) the last item of the sequence is forgotten right before it can be executed:

$h(current\_task(5), 13),$
$o(select(tab(highlighting)), 13),$
$h(current\_task(6), 14),$
$h(task\_forgotten(6), 14).$

In another example, we consider an *expert* user familiar with the "click tools, click options," sequence, but not with the "select tab 'text', put a checkmark on 'use hard-tabs'" sequence. This knowledge is encoded by $\Gamma_2^a$, shown earlier. The prediction of the user's behavior is given by program $\Pi_2^{UI} = \Pi_{STM} \cup \Pi_{TASK} \cup \sigma_{STM} \cup UI_1 \cup \Gamma_2^a$, and it is not

---

[8] These figures are meant to be reasonably realistic, but they are not the result of an accurate study. Obviously, the conclusions of our work hold independently of the particular numbers chosen.

difficult to check that the sequence does not fit in STM:

$$h(current\_task(1), 8),$$
$$h(in\_stm(task(3, select(tab(text)))), 8),$$
$$h(in\_stm(task(4, check(ck(use\_hardtabs)))), 8),$$
$$h(in\_stm(task(5, select(tab(highlighting)))), 8),$$
$$h(in\_stm(task(6, check(ck(detect\_language)))), 8),$$
$$h(task\_forgotten(1), 8).$$

In the final example, an expert user is familiar with both sequences of tasks. The corresponding program is $\Pi_3^{UI} = \Pi_{STM} \cup \Pi_{TASK} \cup \sigma_{STM} \cup UI_1 \cup \Gamma_2$. The answer set of $\Pi_3^{UI}$ shows that the user is predicted to apply chunking, remember the sequence through the end, and complete its execution:

$$h(completed, 15),$$
$$h(in\_stm(task(1, tools\_options)), 15),$$
$$h(in\_stm(task(3, text\_htabs)), 15),$$
$$h(in\_stm(task(5, select(tab(highlighting)))), 15),$$
$$h(in\_stm(task(6, check(ck(detect\_language)))), 15).$$

It is worth stressing that the ASP programs described above are not only directly executable, but the corresponding computation by state-of-the-art ASP systems is quite fast. The answer sets for the examples in this section were each computed in less than a second by CLASP on a computer with i7 CPU, 2.93GHz, 8GB RAM. Scaling is also good: repeating the experiments for values of $\omega$ and $\varepsilon$, respectively, 7 and 100 increased computation time to less than 3 seconds.

## 6 Discussion

*In this paper we have shown that it is indeed possible and important to formalize psychological knowledge that is of qualitative or logical nature, and that ASP is suitable for the task.* The formalization allows analysis and comparison of theories. It also allows one to predict the outcome of experiments, thus making it possible to design better experiments. Various reasons make the formalization of knowledge of this kind challenging. As we hope to have demonstrated, ASP allows one to tackle the challenge, thanks to its ability to deal with common-sense, defaults, uncertainty, non-deterministic choice, recursive definitions, and evolving domains. To highlight the benefits of the formalization of psychological knowledge, it is worth stressing how, earlier in the paper, the availability of the formalization of STM allowed us to point out that the role of more recent symbols in the mechanics of chunking may deserve further investigation.

*We believe it is difficult to find other languages that allow writing a formalization at the level of abstraction of the one shown here, and that are at the same time directly executable.* As a remarkable example, consider how naturally and elegantly inertial and non-inertial fluents coexist in our model, and how the introduction of non-inertial fluents occurs in a fully incremental fashion: for most of Section 4, the model uses only inertial fluents, and one could safely adopt the standard writing of the axioms of inertia.[9] When the need to deal with non-inertial axioms arises later in the section, all that needs to be done is to add

---

[9] In Section 4, we showed directly the more advanced version of the inertia axiom(s) to save space.

the condition "not $noninertial(F)$" to the inertia axioms, and to write suitable definitions of the new fluents. It has been extensively demonstrated in the literature on ASP that the elaboration tolerance shown in this case is not a coincidence, but a key property of the ASP paradigm. Furthermore, direct executability of ASP provides a unique opportunity to bridge the gap between formulation of psychological theories and practical applications, as we have shown in the previous section.

Finally, in this paper we have discussed the interaction with a simple graphical interface, and have barely touched upon related topics such as user expertise. However, we expect that similar techniques to the ones shown here can be used to accurately model the role of expertise and attention (e.g. in particular in critical tasks (McCarley et al. 2002)). We believe that other important theories about the mechanics of the brain, e.g. Long-Term Memory, can be formalized following the approach we presented. The formalization of STM itself could also be made richer, for example by using additive fluents (Lee and Lifschitz 2001) to allow concurrent STM updates, by modeling continuous decay (Chintabathina et al. 2005), and by using intentions (Baral and Gelfond 2005) to represent task sequences.